\pdfoutput=1
\documentclass[11pt]{article}

\usepackage[]{emnlp2021}
\usepackage{times}
\usepackage{latexsym}
\usepackage{subfigure}

\usepackage{times}
\usepackage{latexsym}
\usepackage{microtype}
\usepackage{ wasysym }
\usepackage{tikz}
\usepackage{subfigure}
\usepackage{xcolor,colortbl}
\usepackage{siunitx}
\usepackage{soul}
\usepackage{color}
\DeclareRobustCommand{\hlpink}[1]{{\sethlcolor{pink}\hl{#1}}}
\DeclareRobustCommand{\hlgreen}[1]{{\sethlcolor{green}\hl{#1}}}

\usepackage[T1]{fontenc}
\usepackage[utf8]{inputenc}

\usepackage{microtype}

\title{Predicting Text Readability from Scrolling Interactions}

\author{Sian Gooding \\
 Dept of Computer Science and Technology \\
 University of Cambridge \\
 {\texttt{shg36@cam.ac.uk}} \\\And
  Yevgeni Berzak \\
  MIT BCS \\
  {\texttt{berzak@mit.edu}} \\\AND
  Tony Mak \\
 Google\\
  { \texttt{tonymak@google.com}} \\ \And
  Matt Sharifi \\
 Google\\
  { \texttt{mns@google.com}} \\}

\date{}

\begin{document}
\maketitle

\begin{abstract}
Judging the readability of text has many important applications, for instance when performing text simplification or when sourcing reading material for language learners. In this paper, we present a $518$ participant study which investigates how scrolling behaviour relates to the readability of English texts. We make our dataset publicly available and show that (1) there are statistically significant differences in the way readers interact with text depending on the text level, (2) such measures can be used to predict the readability of text, and (3) the background of a reader impacts their reading interactions and the factors contributing to text difficulty.\footnote{Our dataset is available at \url{https://github.com/siangooding/readability_scroll}}
\end{abstract}

\section{Introduction}
There are multiple attributes of a written text that impact how difficult it is to read. This concept is formally known as the \textit{readability} of text, where readability is defined as the sum total of elements within a text that impact a reader's understanding, reading speed and level of interest \cite{dale1949concept}. Many factors can influence the readability of text, such as the lexical and syntactic complexity, level of conceptual difficulty and style of writing \cite{xia2016text}. For instance, Figure \ref{fig:example_sentences} presents an example sentence (b) that has been rewritten to a more readable format shown in (a).\footnote{Example from the OneStopEnglish dataset by \citet{vajjala2018onestopenglish}.} 

\begin{figure}[t]
\centering
\begin{tabular}{l}
\hline
\rowcolor[HTML]{a9ffac} 
\multicolumn{1}{|l|}{\cellcolor[HTML]{a9ffac}\begin{tabular}[c]{@{}l@{}}(a) A new scientific study says that global\\ warming might make temperatures rise \\ more than people think.\end{tabular}}                                                                   \\ \hline
                                                                                                                                                                                                                                                                      \\ \hline
\rowcolor[HTML]{ffaca9} 
\multicolumn{1}{|l|}{\cellcolor[HTML]{ffaca9}{\color[HTML]{333333} \begin{tabular}[c]{@{}l@{}}(b) Temperature rises resulting from\\ unchecked climate change will be at \\ the severe end of those projected, \\ according to a new scientific study.\end{tabular}}} \\ \hline
\end{tabular}
\caption{Example of two sentences (a) and (b) with differing levels of readability. Sentence (b) has a more sophisticated syntactic structure and advanced vocabulary than (a).}
\label{fig:example_sentences}
\end{figure}

Automatically measuring the readability of text has many useful applications. For example, it is used in text simplification \cite{aluisio2010readability}, when sourcing reading material for language learners \cite{xia2016text}, in ranking search engine content for dyslexic users \cite{10.1145/3173574.3173609} and to ensure critical consumer information is delivered at an appropriate level \cite{10.1145/3290605.3300424}. However, current approaches for measuring readability rely exclusively on linguistic features which do not account for the subjective needs of readers. As a result, traditional readability formulas perform poorly in modeling adult judgements of textual complexity \cite{crossley2017predicting}.

Furthermore, traditional readability approaches do not work well for online content \cite{collins2013enriching}. This is due to systems being highly sensitive to noise; requiring well formed sentences and performing poorly on short passages \cite{collins2014computational}. Linguistically driven techniques are also language specific and require sophisticated models to extract features. For low-resource languages, such tools and models may be unavailable \cite{agic2016multilingual}. 
%Furthermore, in practice there can be difficulties extracting text from online content such as webpages and PDF files \cite{ramakrishnan2012layout} ruling out the use of these features.

Achieving a more inclusive assessment of text readability requires collecting subjective measures on multiple levels (e.g., reading interactions and comprehension-based questions) across different genres and populations. Therefore, alternative approaches must be examined to further the understanding of text difficulty and the multiple factors that influence it. In this paper, we
introduce such an alternative approach, by using crowdsourcing to collect scroll-based interactions whilst participants read texts at differing levels. We present a $518$ participant study, recording the reading interactions of individuals and obtain statistically significant results showing that reading interactions differ dependent on the textual complexity. 

In the following sections, we discuss relevant previous work, outline our data collection and report results from our study. As a preliminary evaluation, we report the statistical significance of reading interactions based on text difficulty, and integrate scroll features into a readability classifier. Finally, we investigate group differences in readability based on the first language of participants. To conclude, in this paper we make the following contributions:
\begin{itemize}
    \item We release a dataset containing the reading interactions and comprehension scores of $518$ participants based on the OneStopEnglishQA dataset \cite{berzak2020starc}.
    \item We show that scroll-based reading interactions can be used to predict the readability of text as well as improve current state-of-the-art approaches. 
    \item We investigate how first language impacts reading interactions and emphasise that the factors contributing to text difficulty vary depending on the target audience.  
\end{itemize}

\section{Related Work}

\subsection*{Text Readability}
Research on assessing the readability of English texts has spanned several decades. The earliest works focused on the construction of readability formulae and metrics \cite{chall1958readability, klare1963measurement, zakaluk1988readability}. These measures rely on shallow textual characteristics such as the number of sentences, average length of sentences and average word length. For example, one of the most well known readability scores is the Flesch-Kincaid score \cite{kincaid1975derivation}. This score takes the average number of words per sentence as well as the average number of syllables as a proxy for syntactic and semantic difficulty. However, there are multiple limitations to readability formulae \cite{collins2014computational}. Firstly, these formulae are based on surface characteristics of text, and ignore deeper levels of text processing known to be important factors of readability. Furthermore, these metrics typically assume the text will contain no noise. As a result of this, a number of studies have demonstrated that these metrics are unreliable for web-based content \cite{si2001statistical, collins2004language, feng2009cognitively}. 

Due to the limitations of traditional formulae, research in readability assessment subsequently shifted towards machine learning (ML) based techniques. These approaches combine a far richer variety of linguistic features using ML classification algorithms and result in far better performance \cite{franccois2012nlp}. Early work on statistical readability assessment demonstrated the improvement of these approaches for readability prediction \cite{si2001statistical, collins2004language}. Subsequent work focused on the addition of appropriate features, for instance, lexical and grammatical \cite{heilman2007combining, heilman2008analysis} as well as discourse-based \cite{pitler2008revisiting,feng2010comparison, graesser2011coh}. Systems specifically designed for non-native audiences have also been developed \cite{feng2010comparison, vajjala2012improving, xia2019text}. Additionally, the use of eye-tracking data from both language learners and native speakers has been shown to improve readability assessment models \cite{gonzalez2018learning}. However, many of the aforementioned models require extensive feature engineering, which not only rely on well-formed content, but also depend on multiple resources such as parsers and word-lists. 

\subsection*{Implicit Feedback Techniques}
Reading on a device, such as a tablet or phone, has predominantly taken the place of traditional formats \cite{qisheng2019browser}. Such devices allow access to implicit user feedback by measuring how a user interacts with the text they read. A key advantage of implicit feedback techniques is that they can unobtrusively obtain information by measuring user interactions with a system \cite{kelly2003implicit}. For instance, \citet{claypool2001implicit} designed a study to capture mouse and keyboard interactions as implicit measures of interest. They measure the correlation of these implicit features with user ratings and find the time spent and amount of scrolling had a positive correlation with interest. \citet{chen2021factors} analyse the factors predictive of English language typing times to investigate effects of linguistic context on language production. Implicit feedback techniques have additionally been shown as useful in information retrieval \cite{golovchinsky1999reading}, ranking summaries \cite{white2002finding} and identifying user preferences \cite{kelly2003implicit}.

In our study, we measure implicit feedback from participant interactions whilst reading. The goal is to produce a set of readability features, using aggregate scroll interactions, that are robust to noise and do not require extensive feature engineering or external resources. 

\begin{figure*}[t]
    \centering
    \begin{minipage}{0.40\textwidth}
        \centering
        \includegraphics[width=5cm]{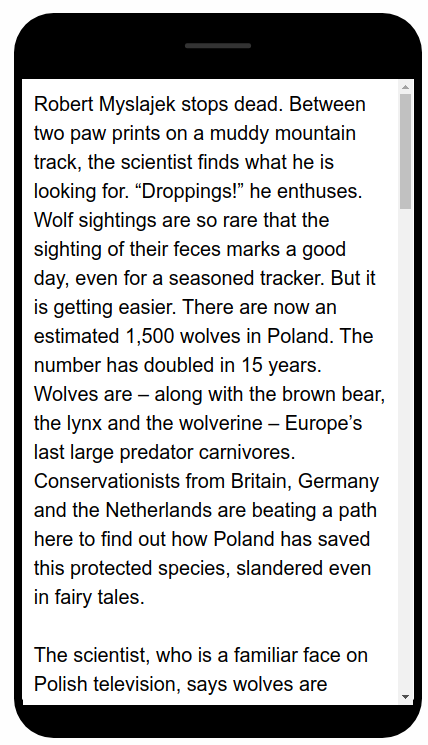}
         \caption{Example of the reading interface presented to participants.}
        \label{fig:interface}
    \end{minipage}\hfill
    \begin{minipage}{0.50\textwidth}
        
\begin{tabular}{l}
\textit{Elementary}                                                                                                                                                                                                                                                                                                                                                                                                                                                  \\ \hline
\rowcolor[HTML]{FFFFFF} 
\multicolumn{1}{|l|}{\cellcolor[HTML]{FFFFFF}\begin{tabular}[c]{@{}l@{}}\small{The brown beer bottle was in the water for 101 years. \hlgreen{A}}\\ \small{\hlgreen{fisherman found it in the Baltic Sea} off the Northern city}\\ \small{of Kiel. \hlpink{Holger von Neuhoff a curator at the museum}}\\ \small{said this bottled message was the oldest he had ever seen.}\end{tabular}}                                                                                                                              \\ \hline
\textit{Advanced}                                                                                                                                                                                                                                                                                                                                                                                                                                                    \\ \hline
\rowcolor[HTML]{FFFFFF} 
\multicolumn{1}{|l|}{\cellcolor[HTML]{FFFFFF} \begin{tabular}[c]{@{}l@{}}\small{The brown beer bottle, which had been in the water for}\\ \small{101 years, \hlgreen{was found in the catch of Konrad Fischer,} a}\\ \small{\hlgreen{a fisherman,} who had been out in the Baltic Sea off of} \\ \small{the northern city of Kiel. \hlpink{Holger von Neuhoff, curator for}} \\ \small{\hlpink{ocean and science at the museum,} said that this bottled}\\ \small{message was the oldest he had come across.}\end{tabular}} \\ \hline
\begin{tabular}[c]{@{}l@{}}\\\small{Q: How was the bottle discovered?}\\ \small{a: A fisherman discovered it}\\ \small{b: It washed up on the shore of Kiel}\\ \small{c: Holger von Neuhoff discovered it}\\ \small{d: A boy discovered it while playing in the sea}\end{tabular}                                                                                                                                                                                                                   
\end{tabular}
         \caption{Example paragraph from \textit{elementary} and \textit{advanced} texts. In the figure, answer spans are highlighted in green and distractor spans in pink.}
  \label{fig:q_example}
    \end{minipage}
\end{figure*}

\section{Data Collection}
We design a controlled reading task and record user reading interactions. We collect responses from participants in the United States and India. This decision was made based on our pilot studies which showed participants from India were much more likely to know English as a second language. The experiment was designed in line with common practises in scroll interaction research \cite{hinckley2002quantitative}. This study represents best efforts in the collection of a novel dataset of reading interactions for text at different levels of readability.

\subsection*{Sampling and Participants}
Participants in the study were recruited via a crowdsourcing platform and were based in either India or the US. A range of background information pertinent to language proficiency was collected using a demographic questionnaire which is available on our dataset repository. This included the self-reported English proficiency, native language, hours spent reading per week and highest level of formal education. The design of the questionnaire was informed following guidelines on judging reading ability \cite{acheson2008new}. English was the first language of $66$\% of participants and a total of $46$ distinct first languages were recorded. The second most frequent first language was Tamil, accounting for $19$\% of participants. 

\subsection*{Materials}
Investigating how readability impacts user interactions required texts at different levels of difficulty. For this, we use the OneStopEnglishQA dataset \cite{berzak2020starc} which has been designed for the evaluation and analysis of reading comprehension in English. The dataset contains $30$ articles from a prior dataset collected for readability assessment and automatic text simplification: OneStopEnglish \cite{vajjala2018onestopenglish}. Each article has been rewritten by teachers to suit three levels of adult ESL learners (\textit{elementary}, \textit{intermediate}, and \textit{advanced}) and are originally from the Guardian newspaper.\footnote{\url{https://www.theguardian.com/uk}} OneStopEnglishQA additionally contains high quality text comprehension questions that are explicitly linked to textual spans. In our study, we use only the \textit{advanced} and \textit{elementary} texts and ask the same questions per article independent of the level. Figure \ref{fig:q_example} illustrates an example of a comprehension question and the relevant textual spans for both an \textit{advanced} and \textit{elementary} paragraph. For each text, we present three comprehension questions that have been strategically chosen to necessitate scrolling. The texts presented to participants are chosen randomly, however a participant will never be shown the same article at differing levels.

\subsection*{Task and Procedure}
The task required participants to read articles and then answer three comprehension questions. Initially, participants were shown instructions as well as a placeholder text with questions before confirming that they were ready to begin. Participants were presented with two texts, one at an \textit{advanced} level and one at an \textit{elementary} level. Texts were shown one at a time in a random order. Participants were immediately given the comprehension questions after reading the text. The instructions stated that participants must read the article and then answer the comprehension questions. However, the text was still accessible during question answering and we record these interactions separately to reading. Additionally, participants were informed that they would be awarded a financial bonus on top of their base rate of pay for each question answered correctly. The decision to award a bonus was to encourage participants to read the text carefully. When presented with an article, participants were not able to progress until at least $90$ seconds had elapsed to encourage engagement and prevent immediate skipping. Once the participant had read the two texts and answered all questions, a demographic questionnaire was presented. 

\subsection*{Implementation}
We used the Qualtrics platform to create the survey and added a custom front end implementation using HTML, CSS and JavaScript. The experiment interface is illustrated in Figure \ref{fig:interface} and was displayed via a browser window. The article text was presented to participants in a restricted window of size $1080$ by $1920$ CSS px, a density-independent measure, based on the dimensions of an average Android device. The font used in the study was sans-serif size $18$pt.

We log an event whenever the participant scrolls on the text. Events are logged every $100ms$ and include a timestamp indicating the elapsed time as well as the scroll y-axis offset in pixels (the vertical distance from the top of the text to the current location).

\begin{table*}[]
\begin{tabular}{lllllllll}
\multicolumn{1}{l|}{Norm. Measures}    &       & \multicolumn{3}{c}{$Elementary$} & \multicolumn{3}{c}{$Advanced$} &       \\
\multicolumn{1}{c|}{($/ length$)}       &    $\times10^{n}$   &  \hphantom{$-$}$\bar{X}$      &  \hphantom{$-$}$\sigma$   &  \hphantom{$-$}$r$          &  \hphantom{$-$}$\bar{X}$      &  \hphantom{$-$}$\sigma$  &  \hphantom{$-$}$r$         & $Sign.$ \\ \hline
\multicolumn{1}{l|}{Read time ($s$)}    &  $\times10^{2}$     &   \hphantom{$-$}$3.4$     &   \hphantom{$-$}$3.8$      &     \hphantom{$-$}$0.10$        &  \hphantom{$-$}$2.1$      &    \hphantom{$-$}$1.9$    & \hphantom{$-$}$0.14$ $\newmoon$   &  $p  < 10^{-18}$     \\
\multicolumn{1}{l|}{Regression num}   & $\times10^{-3}$ & \hphantom{$-$}$5.9$    & \hphantom{$-$}$9.3$     & $-0.09$       & \hphantom{$-$}$7.1$    & \hphantom{$-$}$8.0$    & $-0.03$      & $p  < 10^{-3}$      \\
\multicolumn{1}{l|}{Min speed ($px/ms$)}         & $\times10^{-5}$ & \hphantom{$-$}$4.1$    & \hphantom{$-$}$11$      & $-0.17$ $\newmoon$   & \hphantom{$-$}$3.3$    & \hphantom{$-$}$8.7$    & \hphantom{$-$}$0.01$       &   -    \\
\multicolumn{1}{l|}{Max speed ($px/ms$)}         & $\times10^{-3}$ & \hphantom{$-$}$2.4$    & \hphantom{$-$}$1.7$     & $-0.23$ $\newmoon$   & \hphantom{$-$}$1.7$    & \hphantom{$-$}$1.3$    & $-0.12$      &  $p  < 10^{-12}$     \\
\multicolumn{1}{l|}{Avg speed ($px/ms$)}         & $\times10^{-4}$ & \hphantom{$-$}$7.9$    & \hphantom{$-$}$6.0$     & $-0.35$ $\newmoon$   & \hphantom{$-$}$6.1$    & \hphantom{$-$}$4.6$    & $-0.26$ $\newmoon$  &   $p  < 10^{-14}$    \\
\multicolumn{1}{l|}{Min acc ($px/ms^{2}$)}           & $\times10^{-6}$ & $-5.9$   & \hphantom{$-$}$8.5$     & \hphantom{$-$}$0.12$        & $-4.4$   & \hphantom{$-$}$5.8$    & \hphantom{$-$}$0.08$  &  $p  < 10^{-5}$     \\
\multicolumn{1}{l|}{Max acc ($px/ms^{2}$)}           & $\times10^{-6}$ & $-4.7$   & \hphantom{$-$}$8.7$     & $-0.11$       & $-3.4$   & \hphantom{$-$}$5.8$    & $-0.08$      &  $p  < 10^{-3}$     \\
\multicolumn{1}{l|}{Avg acc ($px/ms^{2}$)}           & $\times10^{-7}$ & $-3.9$   & \hphantom{$-$}$13$      & \hphantom{$-$}$0.07$        & $-2.0$   & \hphantom{$-$}$6.7$    & \hphantom{$-$}$0.07$       &   $p  < 10^{-3}$    \\
\multicolumn{9}{c}{$\newmoon$: $p<0.01$ \hphantom{  }$\bar{X}$: Mean value \hphantom{  }$\sigma$:  Standard deviation}                                                                                        
\end{tabular}
\caption{Interaction measures for $518$ participants across \textit{elementary} and \textit{advanced} texts. Measures have been normalised according to text lengths. The correlation ($r$) of measures with comprehension scores is presented (p-values have been Bonferroni-corrected). The statistical significance of group differences (Sign.) is calculated using a mixed-effects model.}
\label{tab:results_cor}
\end{table*}

\subsection{Preprocessing}
 We collected responses from $600$ participants in total. However, the dataset was initially preprocessed to mark entries where the participant had not sufficiently engaged with reading the text. Given the size of the screen and text lengths, scrolling was necessary in order to read the text. We only included participants in our analysis if their scroll pattern indicated that they had attempted to read the article. If no scrolling had been logged, or the participant had not reached the half way point, the entry was not included in the analysis. Removing these entries resulted in $518$ participants and $1036$ articles marked as read.  

\subsection*{Interaction Measures}
As a preliminary analysis, a range of interaction metrics were extracted from the scrolling behaviour of participants. The computation of these follow the standard of those used in prior scroll research \cite{hinckley2002quantitative}. These measures are outlined below.

 \textsc{Elapsed time:}
The total reading time is recorded to produce a \textit{read time} in seconds.\\

\textsc{Speed:}
The scroll speed ($s$) is calculated for each scroll interaction using $s = d / t$, where $d$ represents the distance in pixels and $t$ the time taken in $ms$. The average, minimum and maximum scroll speed are calculated. \\

\textsc{Acceleration:} 
Scrolling acceleration ($a$) is computed for each interaction. This is calculated with the following formula:  $a = (v - u) / t$ where $v$ represents the final scroll speed, $u$ the initial speed and $t$ denotes the time taken in $ms$. The average, minimum and maximum scroll acceleration are calculated. \\

\textsc{Text regressions:}
Scrolling typically takes place in a linear vertical fashion. Whilst reading, areas of text may require re-covering resulting in upward scrolling actions. We count the number of times the participant scrolled upwards to recover areas of text.

\section{Results}
\subsection*{Readability Measures}
Table \ref{tab:results_cor} shows the mean values ($\bar{X}$) and the standard deviation ($\sigma$) of reading interactions normalised by text length. We additionally present the correlation ($r$) of these with the comprehension scores of participants. P-values have been Bonferroni-corrected to account for multiple test conditions. The results show that three measures correlate significantly with participant scores on \textit{elementary} texts and two for \textit{advanced}. For the \textit{elementary} articles, all speed measures correlate significantly with participant scores. The negative correlation shows that the slower the speeds, the higher the subsequent score. For \text{advanced} texts, the average scroll speed also negatively correlates with score. Whereas the time taken to read the article positively correlates with the subsequent score.  

When considering the mean values across interactions, we see that all speed and acceleration measures are lower for \textit{advanced} texts. This finding shows that, on average, the speed and acceleration of scrolling is slower on texts that are more difficult. Additionally, the number of regressions is larger for \textit{advanced} texts than for \textit{beginner}. Therefore, participants were more likely to recover areas of text when it is at a higher level. Finally, the standard deviation is larger for all measures on the \textit{elementary} texts. This implies that there is more variance in reading interaction styles when the text is at a lower level.  

We consider whether reading interactions differ significantly depending on the level of the text. We calculate significance using Satterthwaite’s method \cite{kuznetsova2017lmertest}, applied to a mixed-effects model that treats participants and texts as crossed random effects.\footnote{Using R formula notation, the model is:
$measure \sim readability + (readability|participant) + (readability|text)$. Tests were performed using the lme4 and lmerTest R packages by \citet{bates2014fitting}.} All measures are found to be statistically significant apart from the minimum speed. The most significant measure ($p < 10^{-18}$), is the normalised time a participant spent on the text. Counter-intuitively, the time taken, on average, is longer for the \textit{elementary} texts than for \textit{advanced}. However, when we consider proficiency groups, participants with lower proficiency levels took longer on \textit{elementary} texts compared to \textit{advanced}. This suggests that for readers at a low level the text is too difficult to engage with, resulting in the skipping of content. All acceleration measures, the maximum and average speed as well as the number of regressions differ significantly depending on whether text is \textit{advanced} or \textit{elementary}. These findings support the case that there are different reading interactions for texts depending on their complexity. 

\subsection*{Predicting Readability}
\begin{table*}[t]
\centering
\begin{tabular}{l|llll}
\textit{System} & \textit{$N$} & \textit{Precision} & \textit{Recall} & \textit{F-Score} \\ \hline
Scroll only    & 12                   & 0.80               & 0.78            & 0.77             \\
\textsc{Vajjala-2018}    & 155                   & 0.88               & 0.85            & 0.84             \\
\textsc{Vajjala-2018} + \textit{Scroll}    & 160                   & 0.92               & 0.88            & 0.88             \\
 \hline
Baseline (length + LR)    & 6                     & 0.91               & 0.87            & 0.88             \\ 
Baseline + \textit{Traditional}     & 15                    & 0.92               & 0.90            & 0.89             \\
Baseline + \textit{Discourse}     & 24                    & 0.81               & 0.82            & 0.80             \\
Baseline + \textit{Syntactic}     & 24                    & 0.87               & 0.85            & 0.84             \\
Baseline + \textit{Psycholinguistic}           & 12                    & 0.87               & 0.88            & 0.87             \\
Baseline + \textit{Scroll}        & 18                    & \textbf{0.98}      & \textbf{0.97}   & \textbf{0.96}   
\end{tabular}
\caption{Table showing the total number of features per system ($N$) as well as the precision, recall and f-score of models trained using 10-fold cross-validation on the OneStopQA dataset.} 
\label{tab:scroll_prediction_sota}
\end{table*}

We perform experiments to investigate whether scroll-based reading interactions can be used to predict the level of a given text. As the texts in our dataset have been read by multiple participants, we produce features by taking the average of the statistically significant scroll measures (as displayed under Sign. in Table \ref{tab:results_cor}). Prior research has shown that simple methods of combining group interactions, such as averaging or majority voting, can be highly robust \cite{genre2013combining, clemen1989combining, ertekin2012learning}. We produce two sets of scroll features using the number of regressions, the max and average scroll speed and the max, min and average scroll acceleration. One set is normalised by text length (Normalised) and the other not ($\lnot$ Normalised). We then train a support vector machine (SVM) to predict whether a text is \textit{advanced} or \textit{elementary} using these features. This classifier was chosen as it has been shown to consistently yield better results compared to other statistical models when predicting text readability \cite{kate2010learning}. All reported results are obtained using stratified 10-fold cross-validation.

Table \ref{tab:scroll_prediction} presents the f-score of the resulting models. To the best of our knowledge, this is the first work aiming to predict readability using scroll-based features. The best result ($0.77$) is achieved when using both sets of scroll features. The $\lnot$ Normalised feature set performs better than Normalised. This is likely due to the fact that these features contain signal on the length of texts. 

\begin{table}[t]
\centering
\begin{tabular}{l|ll}
\textit{Scroll features} & \textit{F-Score} \\ \hline
All ($\lnot$Norm + Norm)         & \textbf{0.77}  \\ 
$\lnot$ Normalised    & 0.64   \\
\hphantom{$\lnot$} Normalised   &   0.61   \\
\end{tabular}
\caption{Readability prediction using aggregate interaction features.} 
\label{tab:scroll_prediction}
\end{table}

Interestingly, filtering interactions by specific sub-groups can produce better scores. The best results are achieved, using both sets of scroll measures, when we only include the interactions of participants aged 25-34 -- resulting in an f-score of $0.81$. There are known differences in computer interaction styles based on age \cite{schneider2008investigation}. It therefore follows that aggregate measures from an audience with a more consistent interaction style would result in better features. 

The use of scroll features alone produces an f-score of $0.77$. We compare our result to a state-of-the-art readability system by \citet{vajjala2018onestopenglish} referred to as \textsc{Vajjala-2018}. This system uses a multilayer perceptron classifier and has been shown to outperform BERT-based approaches on the OneStopEnglish dataset \cite{martinc2019supervised}. The system relies on 155 hand-crafted features which are grouped into six categories: traditional metrics, word features, psycholinguistic, lexical richness, syntactic and discourse features. The \textsc{Vajjala-2018} system outperforms the use of scroll features alone when trained to predict if texts are \textit{advanced} or \textit{elementary} on our dataset. However, we are able to improve the f-score of this state-of-the-art system by $4\%$ when adding aggregate scroll features. This is an interesting result as it shows that 1) these features are highly complementary to existing readability systems and 2) these features represent an aspect of textual complexity not covered by the current $155$ features. This may be due to scroll-based features capturing a notion of conceptual complexity which plays a vital role in text understanding and maintaining a reader’s interest \cite{vstajner2020shallow}. 
 
 We also investigate how scroll-interaction features compare to classical readability feature sets (an overview of feature sets is included in Appendix A). To do this, we initially create a highly competitive baseline by training an SVM using the length of text and measures of lexical richness. Due to the nature of the OneStopEnglish dataset, length is an extremely informative feature. This is because, on average, the word length for \textit{advanced} texts ($915$) is almost always higher than for the simplified \textit{elementary} texts ($599$). Vocabulary knowledge, including lexical diversity and richness, are principal components in reading comprehension \cite{collins2014computational}. The importance of lexical richness has been investigated from the perspective of second language acquisition \cite{lu2012relationship}, as well as in readability systems \citep{vajjala2012improving, vajjala2018onestopenglish, xia2019text}.  We opted to use features in our baseline system that were highly informative but would not require extensive text processing. A key advantage of lexical richness measures is that they are a function of the number of word types (T) and the total text length (N). Therefore, they can be calculated quickly and are robust to noisy or broken text. The specific measures of lexical richness we include are type-token ratio (TTR), which is the ratio of the number of unique word tokens to the total number of word tokens in a text, Root TTR $(T/\sqrt{N})$, Corrected TTR $(T/\sqrt{2N})$, Bilogarithmic TTR ($Log T/Log N$) and Uber Index ($Log^{2}T/Log( N/T)$). The resulting baseline approach outperforms the \textsc{Vajjala-2018} system. However, it should be noted that the \textsc{Vajjala-2018} system was originally designed to differentiate between \textit{elementary}, \textit{intermediate} and \textit{advanced} texts which is a more nuanced and difficult task. 
 
  By performing ablation tests, we see how scroll-interaction features compare to classical sets of readability features when combined with a strong baseline system (Length + LR). We find that adding sets of discourse, syntactic or psycholinguistic features degrades classifier performance. The addition of traditional measures, such as Flesch-Kincaid score, produces a small improvement. However, adding aggregate scroll features produces a marked improvement on the baseline resulting in an f-score of $0.96$. The addition of scroll-interaction features improves performance beyond all other classical feature combinations. A key advantage to this approach is that the result is achieved with only $18$ features which are highly robust to noisy text. 

\subsection*{Subjective Readability}
Readability approaches typically produce an objective numerical score which often corresponds to a suggested level \cite{10.2307/20205195}. In our previous experiments, we predict the readability of text as defined by such preordained levels (i.e., \textit{elementary} or \textit{\textit{advanced}}). However, the readability of text can also be defined in a more subjective and idiosyncratic manner. Such techniques have been referred to as \textit{levelling}, and are similar to readability in that they determine difficulty but are subjective \cite{clay1991becoming}. Levelling integrates a reader's background and experience with objective readability by understanding what contributes to readability for differing audiences. In this section, we investigate what reading-interactions could tell us about an individual's text comprehension or L1. 

\begin{figure}[t]
\centering
\includegraphics[width=5cm]{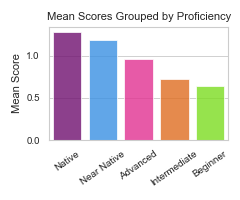}
\caption{Bar chart showing the average scores (out of three) for texts across proficiency levels.}
\label{fig:mean_scores}
\end{figure}

In our study, we record comprehension scores to evaluate the understanding and readability of text for individuals. Participants are asked three questions per article assessing the reader's understanding of what they have read. Figure \ref{fig:mean_scores} shows the average score of participants grouped by their self-reported proficiency. We see that the average score decreases in line with the reported proficiency. %The average score for \textit{elementary} texts was higher ($1.15$) than that of \textit{advanced} ($1.08$). 

%When measuring the significance of correlations p-values are Bonferroni-corrected to account for multiple test conditions. 
In the fashion of levelling, we investigate the correlation of features with the comprehension scores of audiences. We observe whether the feature correlations, at both the text and interaction level, vary depending on the first language (L1) of the group. We focus on English (n = $350$) and Tamil (n = $101$). Tamil was chosen as it was the second most represented L1 after English.

We find there is a statistically significant association between the average scroll speed and scores of participants for both Tamil and English ($p < 10^{-4}$). The correlation is negative, indicating that the faster the average speed of reading, the lower the subsequent score. When considering the English L1 audience, the only statistically significant correlation with score was the average reading speed. However, for Tamil readers there were three statistically significant correlates with ($p < 10^{-4}$), in order of strength these were: the mean age of acquisition for vocabulary in the text (AoA) as reported by \citet{kuperman2012age}; the average reading speed and the mean AoA of word lemmas. The negative correlation shows that the higher the AoA of vocabulary, the lower the score for this group. This finding supports prior work emphasising the importance of AoA as a factor when simplifying texts for second language learners \cite{crossley2007discriminating}. 

\begin{figure}[t]
    \centering
    \includegraphics[width=6cm]{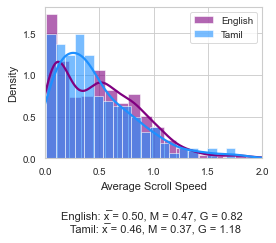}
    \vspace{2mm}
      \includegraphics[width=6cm]{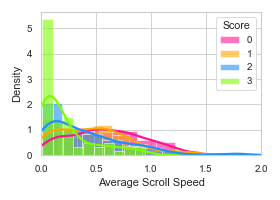}
    \caption{Histograms showing the density of average scroll speed for Tamil and English as well as across participant scores.}
    \label{fig:avg_speed_distribution}
\end{figure}

The average scrolling speed correlates significantly with scores for both English and Tamil L1 audiences. We investigate whether the distribution of average reading speeds differs based on the reader group. Figure \ref{fig:avg_speed_distribution} presents a histogram of the average reading speeds. In both groups, we observe a positively right skewed distribution, as speed is a vector quantity and a range boundary occurs at 0. When comparing speed distributions, the mean and median speeds are higher for English ($0.50$, $0.47$) than Tamil ($0.46$, $0.37$). The higher median and mean measures for English interactions are likely due to a higher familiarity yielding a faster reading speed. The L1 English group additionally achieve a higher score on average ($1.18$) than the Tamil audience ($0.98$), despite the faster scroll speed.

%We calculate the Fisher-Pearson coefficient to measure the skew of distributions. There is a smaller skew in the average reading speed for English compared to Tamil ($0.82$ and $1.18$ respectively). The larger skew of values is likely due to Tamil containing a wider range of proficiency backgrounds, as when we control for participants with a score above 2, this skew reduces to $0.30$. This supports the case that a reader's first language alters their textual interactions.
In Figure \ref{fig:avg_speed_distribution}, we consider how the distribution of average reading speeds varies for participants according to the score they attained. The higher the score, on average, the lower the scroll (reading) speed. For the lower scores, we see that the distribution has a wider spread. According to the self-reported proficiency ratings, the \textit{advanced} texts would likely have been too difficult for a proportion of participants. A higher scrolling speed indicates that a reader is skipping content without properly reading, perhaps due to the level being too high for the reader to sufficiently engage with. Finally, there is a statistically significant positive correlation between the average reading speed of a participant and their reported proficiency ($p < 10^{-4}$), further supporting the notion that reading interactions vary based on ability. 

Being able to understand how on-device reading interactions vary according to an individuals' L1 and comprehension is incredibly useful. Such information could be used in text simplification systems, or in a `levelling' manner to match the appropriate level of text to a given reader. Such applications are especially useful for individuals who are learning a language. 
\\
\section{Conclusions}
To conclude, the use of scroll features for judging readability has numerous benefits. Such measures are language agnostic, unobtrusive and are robust to noisy text. Furthermore, implicit user feedback allows an insight into readability at an individual level, thereby allowing for a more inclusive and personalisable assessment. We present a $518$ participant study to investigate the impact of text readability on reading interactions. In this paper, we confirm that there are statistically significant differences in the way that readers interact with \textit{advanced} and \textit{elementary} texts, and that the comprehension scores of individuals correlate with specific measures of scrolling interaction. We demonstrate that, even with a simple model, aggregate scroll interactions can be used to predict readability. Moreover, we show that individual scroll behaviour can provide an insight into the subjective readability for an individual. Finally, we improve a state-of-the-art readability classifier with the integration of scroll-interaction features, demonstrating that interaction features are highly complementary to traditional linguistic approaches. In future work, we will focus on investigating which aspects of readability scroll-based measures index.

 %Furthermore, the dataset we release with our work contains information on how text interactions differ when an individual is answering comprehension questions vs reading. We aim to investigate whether information retrieval techniques for question answering vary depending on the audience background.  

 % Also could investigate how features like range are informative for readability. Plus, how this works for more levels of text. Could do a short paper on predicting proficiency based on scroll behaviour as follow up too. 
\bibliography{anthology, custom}
\bibliographystyle{acl_natbib}

\section{Appendix}
\subsection*{A - Traditional Readability Feature Sets}
The readability feature sets for the OneStopEnglish dataset were provided by (Vajjala and Lučić., 2018)\footnote{\url{ https://zenodo.org/record/1219041}}. Please refer to their work for further details. An overview of features is provided: 

\subsubsection*{Traditional}
\begin{itemize}
\item Avg. number of characters per word
\item Avg. number of syllables per word
\item Avg. sentence length
\item Flesch-Kincaid score 
\item Coleman-Liau readability formula 
\item SMOG grade 
\end{itemize}

\subsubsection*{Discourse}
\begin{itemize}
    \item Word overlap features - content word, noun, stem and argument overlap at local (between adjacent sentences) and global (between any two sentences in a text) levels.
    \item Entity transitions - features based on the transitions between the syntactic roles of entities. 
    \item Co-reference chains - features based on noun phrases, nouns and pronouns and determiner usage.
    \item Num. of referential expressions.
    \item Num. of discourse and non-discourse connectives and all connectives
\end{itemize}
\subsubsection*{Syntactic}
\begin{itemize}
    \item Num. NPs per sentence (NumNP) 
    \item Num. VPs per sentence (NumVP) 
    \item Num. PPs per sentence (NumPP)) 
    \item Avg. length of a NP (NPSize)
    \item Num. Dependent Clauses per sentence (NumDC) 
    \item Num. Complex-T units per sentence (NumCT)
    \item Num. Co-ordinate Phrases per sentence (CoOrd) 
    \item Num. SBARs per sentence (NumSBAR)
    \item Avg. Parse Tree Height (TreeHeight)
    \item Avg. length of a VP (VPSize)
    \item Avg. length of a PP (PPSize)
    \item Mean length of clause (MLC)
    \item Mean length of a sentence (MLS)
    \item Mean length of T-unit (MLT)
    \item Num. of Clauses per Sentence (C/S) 
    \item Num. of T-Units per sentence (T/S)
    \item Num. of Clauses per T-unit (C/T)
    \item Num. of Complex-T-Units per T-unit (CT/T) 
    \item Dependent Clause to Clause Ratio (DC/C)
    \item Dependent Clause to T-unit Ratio (DC/T)
    \item Co-ordinate Phrases per Clause (CP/C)
    \item Co-ordinate Phrases per T-unit (CP/T) 
    \item Complex Nominals per Clause (CN/C) – Complex Nominals per T-unit (CN/T) 
    \item Verb phrases per T-unit (VP/T)
\end{itemize}

\subsubsection*{Psycholinguistic}
Norms from MRC which were compiled by Gilhooly and Logie (1980) for 1903 unique words including:
 \begin{itemize}
     \item Avg. word age of acquisition
     \item Avg. word Familiarity 
     \item Avg. word concreteness
     \item Avg. word imagability
     \item Avg. word meaningfulness 
 \end{itemize}
\end{document}